\documentclass[10pt,twocolumn,twoside] {IEEEtran}

\usepackage{caption}
\usepackage{setspace}
\usepackage{lineno,hyperref}
\usepackage{times}
\usepackage{epsfig}
\usepackage{graphicx,float}
\usepackage{amsmath}
\usepackage{amssymb}
\usepackage{subfigure}
\usepackage{multirow}
\usepackage{booktabs}
\usepackage{capt-of}
\usepackage[justification=centering]{caption}
              
\usepackage{color}

\DeclareMathOperator*{\argmax}{arg\,max} %

\begin{document}

\title{Security Event Recognition for Visual Surveillance} 

\author{Michael Ying Yang$^*$,~\IEEEmembership{Senior Member,~IEEE,}
        Wentong Liao,
       Chun Yang,
			Yanpeng Cao,~\IEEEmembership{Member,~IEEE~and}
			Bodo Rosenhahn~\IEEEmembership{Member,~IEEE}
\thanks{Michael Ying Yang is with Scene Understanding Group, University of Twente, The Netherlands {\tt\small (e-mail: michael.yang@utwnte.nl)}.}
\thanks{Wentong Liao, Chun Yang, Bodo Rosenhahn are with Leibniz University Hanover, Germany {\tt\small (e-mail: liao@tnt.uni-hannover.de)}.}
\thanks{Yanpeng Cao is with Zhejiang University, Hangzhou 310027, China.}

}

\maketitle

\begin{abstract}
With rapidly increasing deployment of surveillance cameras, the reliable methods for automatically analyzing the surveillance video and recognizing special events are demanded by different practical applications.
This paper proposes a novel effective framework for security event analysis in surveillance videos. 
First, convolutional neural network (CNN) framework is used to detect objects of interest in the given videos. 
Second, the owners of the objects are recognized and monitored in real-time as well.
If anyone moves any object, this person will be verified whether he/she is its owner.
If not, this event will be further analyzed and distinguished between two different scenes: moving the object away or stealing it.
To validate the proposed approach, a new video dataset consisting of various scenarios is constructed for more complex tasks.
For comparison purpose, the experiments are also carried out on the benchmark databases related to the task on abandoned luggage detection.
The experimental results show that the proposed approach outperforms the state-of-the-art methods and effective in recognizing complex security events.
\end{abstract}


\section{Introduction}
\label{introduction}

\IEEEPARstart{S}{e}curity at public place has always been one of the most important social topics.
With rapidly increasing deployment of surveillance cameras in different scenes, tones of video data need to be analyzed in every second.
Conventional surveillance systems almost completely rely on security workers to keep watching the surveillance monitors and recognize the suspected person and activities. It's a low efficient and reliable but costly. 
Therefore, the reliable methods for automatically analyzing the surveillance videos and reporting special events are demanded by different practical applications, such as security monitoring~\cite{collins2000system,liao_gp_2015}, traffic controlling~\cite{wang2009unsupervised,LiaoRY15}, crime prevention \cite{piza2016crime} etc.
Due to their large market and practical impact, much attention has been drawn in computer vision community for decades \cite{fan2011modeling,liao2008localized,evangelio2011detection,fan2013relative,lin2015abandoned}.
The task of security event analysis refers to suspicious object detection and anomaly detection in given videos.

Since the object type of category occurring in surveillance scene is unexpected, traditional methods ignore the object type and use foreground/background extraction techniques to identify static foregrounds regions as suspicious object candidates. 
However, object type provides very important information for video event analysis. 
For instance, a black luggage is more suspicious than a pink wallet which has been left on the floor in an airport hall. 
Only detecting static items is insufficient to deeply and correctly analyze such complicated circumstance.
The main reason that the previous works only focus on abandoned/left-luggage detection is the imperfect object detector which can only detect limited kinds of object categories with  unsatisfied accuracy.
%
%
In recent years, convolutional neural networks (CNNs) are driving advances in computer vision, such as image classification \cite{krizhevsky2012imagenet}, detection \cite{girshick14CVPR,ren2015faster,LiuLWYY16,GirshickDDM16}, semantic segmentation \cite{long_shelhamer_fcn_2015,MustikovelaYR16}, pose estimation \cite{toshev_pose_2014,KrullBMYGR15}.
CNNs have shown remarkable performance in the large-scale visual recognition challenge (ILSVRC2012) \cite{russakovsky2015imagenet}. 
The success of CNNs is attributed to their ability to learn rich feature representations as opposed to hand-designed features used in traditional image classification methods. 
Therefore, it is a good choice to use deep learning methods to detect object type in the task of security event recognition.

\begin{figure}[http]
\begin{center}
\fbox{
\includegraphics[width=0.5\textwidth]{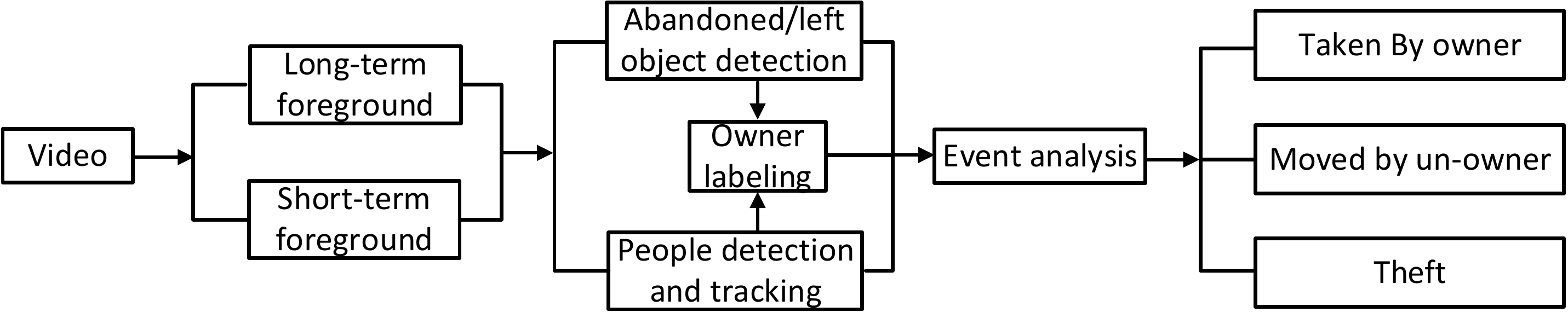}
}
\end{center}
\caption{Flowchart of our framework.}
\label{fig:flowchart}
\end{figure}

Our goal in this work is to detect abandoned objects and then analyze the latter events related to them: its owner is taking it, or someone else is moving it to somewhere, or stealing it? These three security events are the most often occurring circumstances in our daily life.
In this paper, CNN framework is used for object detection and verification. 
Because the previous works only focus on left object detection, appropriate benchmark dataset is missing for more complicated tasks. 
Therefore, we construct a new video event dataset: Security Event Recognition Dataset(SERD) containing various scenarios within real-world environment.
We evaluate our method on the benchmark PETS2006. 
Besides our framework are evaluated on our dataset SERD for further more complicated tasks.
Quantitative and qualitative comparisons with ground truth show that the proposed framework is effective for security event detection.

\section{Methodology}
\label{Sec:method}
Our framework is described by the key components of person and object detection, ownership labeling and security event analysis. 
An overlook of our framework is illustrated in Fig.~\ref{fig:flowchart}.
In the following subsections, each component is discussed in details.

\subsection{Background Model}
\label{bg_fg}
Static is the most obvious character of abandoned objects. 
Thus, our framework apply dual-background model to detect static regions as candidates of abandonment.
The background is divided into long-term which is used for detecting static foreground objects, and short-term one for moving objects.
Long-term background model at time point $t$ is denoted as $\mathbf{BG}_L^t$ and the short-term one is $\mathbf{BG}_S^t$. We denote $\mathbf{F}_L^t$  as binary foreground image obtained via $\mathbf{BG}_L^t$, and $\mathbf{F}_S^t$ via $\mathbf{BG}_L^t$.
 
The background model proposed in~\cite{russell2006minimum} is utilized in our framework because of its high effectiveness and efficiency. 
In our application, 20 frames of each 50th frame are sampled for updating long-term background model and each 3th frame for short-term background model. 
With frame rate of 25Hz, the long-term background completely updates in each 40 seconds and the short-term background updates each 2 seconds.

\subsection{Person and Object Detection}
\label{detection}
In recent years, deep learning based algorithms have shown great power in object detection and classification tasks~\cite{russakovsky2015imagenet,redmon2016you,ren2015faster}.
Considering the trade-off between "real-time" capability and accuracy, the faster region proposal convolution neural network (FrRCNN)~\cite{ren2015faster} is applied in our framework.

According to the ownership relationship, all objects of interest are divided into background objects and foreground objects.
First, the FrRCNN is used to detect objects from the learned initial long-term background RGB image. 
These detected objects are registered in $\mathbf{BO}=\{BO_1,\dots,BO_{N_B}\}$, which indicates that these objects belong to the background. 

As mentioned before, objects abandonment  and security events relate to static objects.
There are three states for object: If a object presents in the long-term foreground but not in the short-term foreground, it is static.
If it presents in both foreground masks, it is moving. If an object has ever presented in the foregrounds but disappears from both of the foregrounds later, it means that it is in static for a very long time.

Therefore, an $XOR$ operation is conducted between $\mathbf{F}_L^t$ and $\mathbf{F}_S^t$ to get the static foreground regions as candidates of static objects.
Then, FrRCNN is applied to detect objects of interest within those static regions.
The FrRCNN is only applied to detect objects within the foreground regions instead of the whole image to reduce computation, which is important for real-time application.
All the objects detected in this step are static objects and registered in a list of $\mathbf{SO}=\{SO_1,\dots,$
$SO_{N_O}\}$, where $SO_i$ encodes the information of a single object category, bounding box, and its features which will be discussed in Sec.~\ref{re_id}.
We utilize the same FrRCNN which is mentioned above to detect Persons on each RGB frame and denote them as $\mathbf{P}=\{P_1,...P_{N_p}\}$. $P_i$ includes the extracted features of the detected person.
Subsequently, the real-time tracking algorithm proposed by Bewley et al.~\cite{bewley2016simple} is utilized in our framework for tracking.
The tracing information of each person is denoted as $T_i$.

\subsection{Ownership Labeling and Abandoning Detection}
\label{ownership}

The owner of an objects is one of the most important information to make sure whether an object is abandoned or just left provisionally.
It is also the crucial cue to analyze the security events, such as theft.
Thus, to identify the owner, we compute the average distance between $SO_i$ and each person's trace $T_i$ over time. 
The person with smallest distance to $SO_i$ is labeled as the owner and denoted as $OP_i$.
Because the shot-term background is updated in each 2 seconds in our work, only the section of each trace from $T_i^{t-2s}$ to $T_i^t$ is considered, whereas $t$ is the time point of $SO_i$ being detected.

\begin{figure}[htp]
\begin{center}
\includegraphics[width=0.6\linewidth]{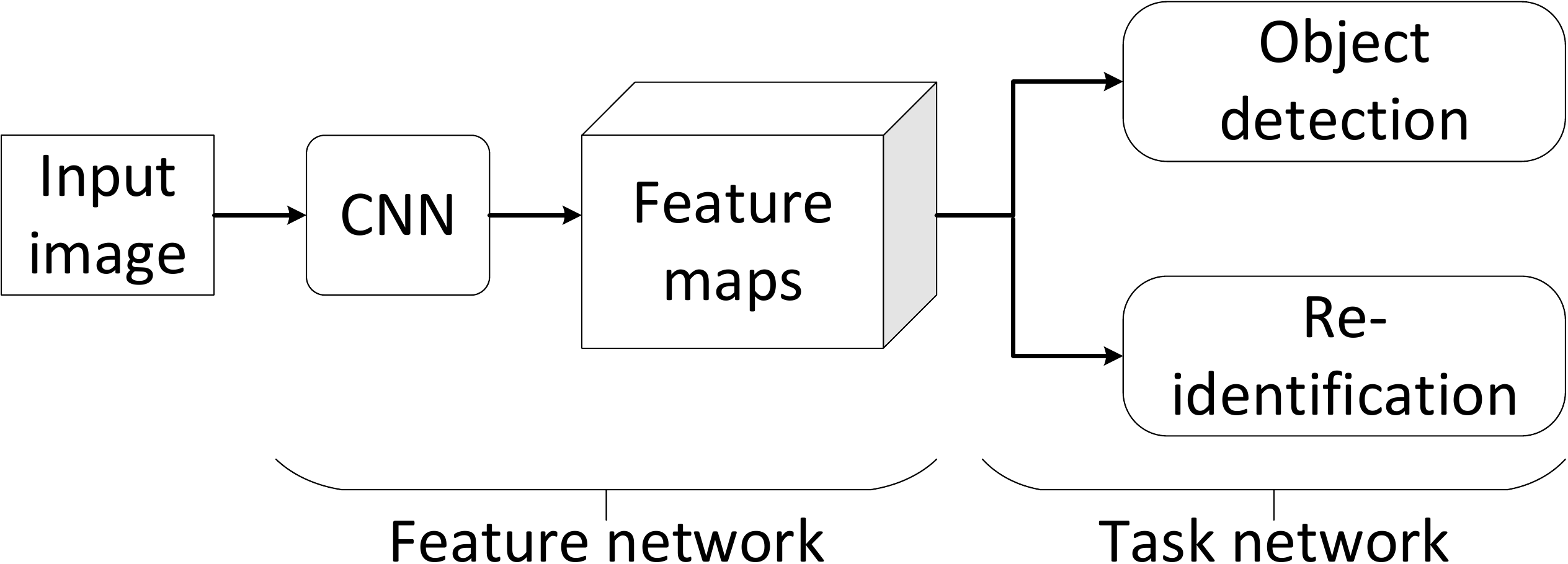}
\end{center}
\caption{A deep CNN can be divided into two parts: the first part is used to extract deep features and dubbed as feature network; the second part is used to finish specific task, such as object detection, and dubbed as task network. Different tasks can used the same feature networks to save training cost, in particularly the tasks are similar to each other.}
\label{fig:shared_cnn}
\end{figure}

\begin{figure*}
\begin{center}
\fbox{
\begin{minipage}{1.0\linewidth}
\subfigure[People detection]{
\label{sugfig:come_in}
\includegraphics[width=0.23\linewidth]{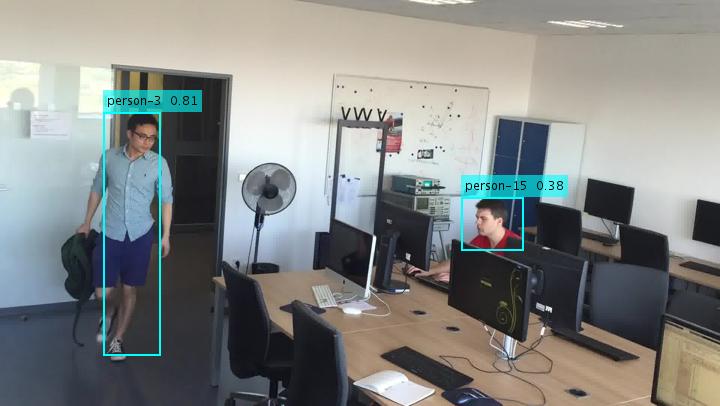}}
\subfigure[Object and owner detection]{
\label{subfig:obj_table}
\includegraphics[width=0.23\linewidth]{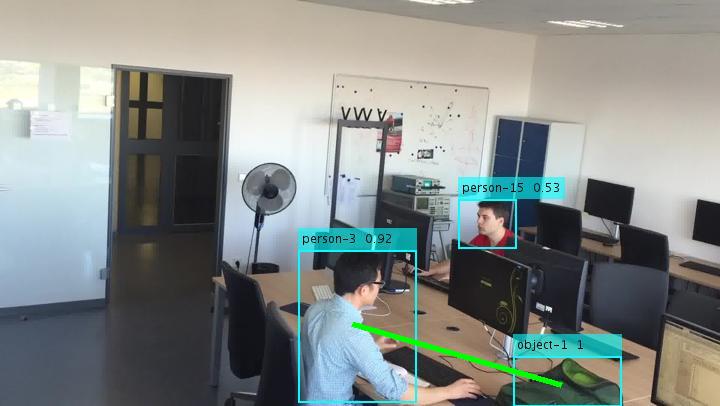}}
\subfigure[New people detection]{
\label{subfig:new_people}
\includegraphics[width=0.23\linewidth]{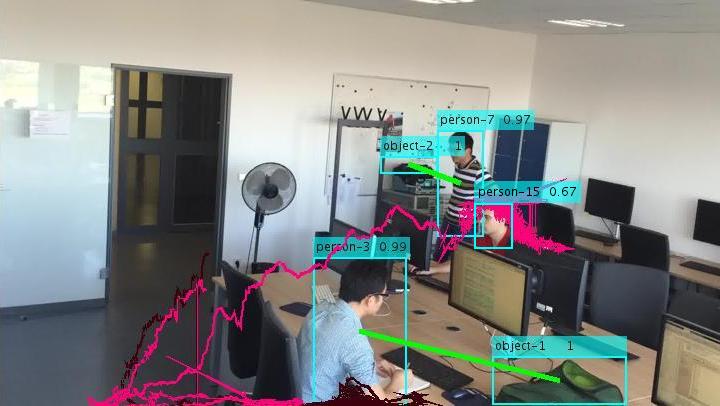}}
\subfigure[Owner is leaving]{
\label{subfig:new_object}
\includegraphics[width=0.23\linewidth]{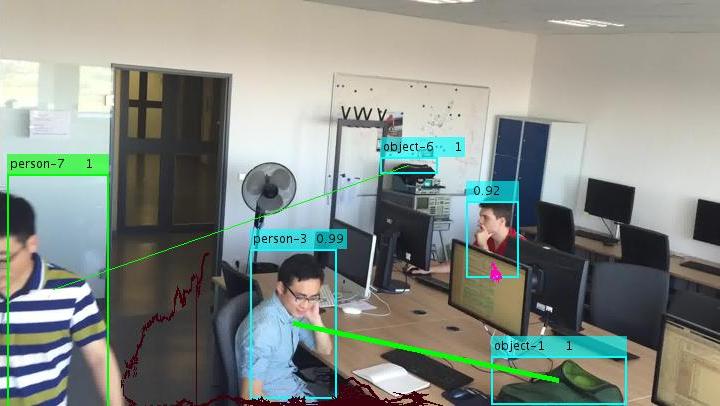}}\\
\subfigure[Bag is taken by owner]{
\label{subfig:takes_his_bag}
\includegraphics[width=0.23\linewidth]{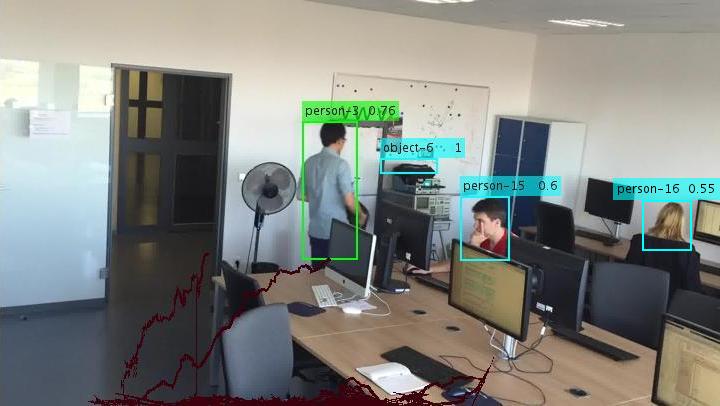}}
\subfigure[Person switches the bag]{
\label{subfig:exchange}
\includegraphics[width=0.23\linewidth]{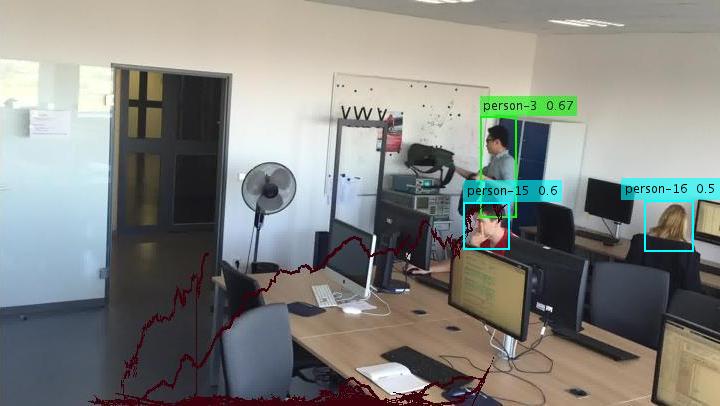}}
\subfigure[Alarm is triggered]{
\includegraphics[width=0.23\linewidth]{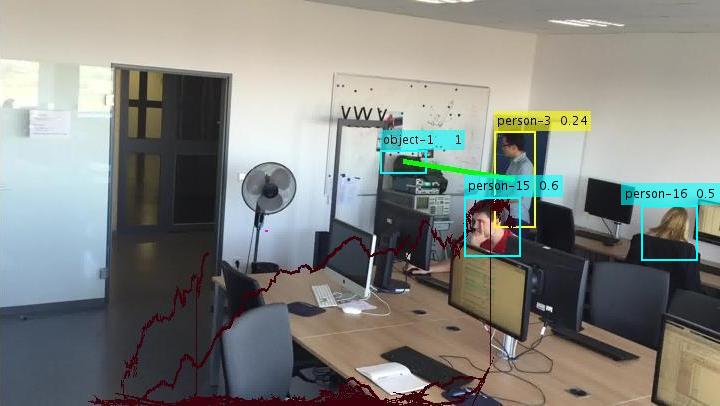}}
\subfigure[Stealing detection]{
\label{subfig:steal2}
\includegraphics[width=0.23\linewidth]{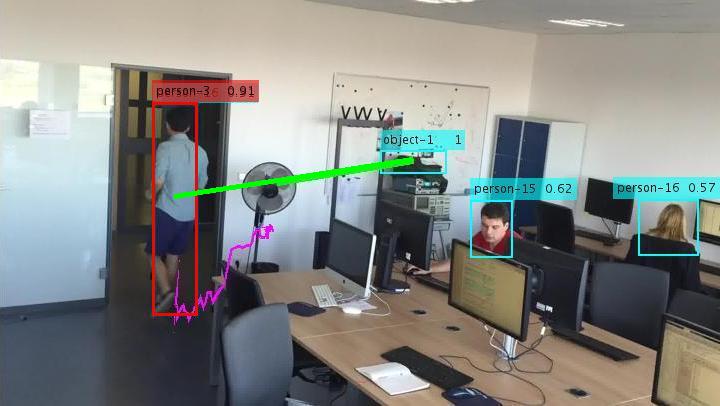}}
\end{minipage}
}
\end{center}
   \caption{An example of experimental results on SERD. A man comes into the room (a). Then he left his bag on the table and begins to work(b). He is labeled as the owner but the bag is not labeled as abandoned object. Another man left his before the white board (c) and left the scene (d). He is labeled as the owner and the bag is labeled as abandoned object. The owner takes his bag away without alarm (e). He switches the bag by his (f), and a warning is issued that the bag does not belongs to him and he is labeled as the owner of the substitute objects (g). He is recognized as a theft when he is leaving (h).}
\label{fig:stu_lab2}
\end{figure*}

It is costly and unnecessary to watch over all objects appearing in the surveillance scene.
Security events in public scenes relate to abandoned objects normally.
Therefore, abandonment should be detected reliably.
The basic rule for abandoned object detection are originally defined by PETS2006:
From temporal aspect, if an object is unattended move his bag in $30$ seconds, the bag is declared as an abandonment;
From the spatial aspect, an object is defines as abandonment if there is not owner within 3 meters.
However, in practice the owner may stay in the scene for a very long time without touching his object. For instance, in the public rest area of a library, a student who wants a break put his bag on a table and then go to a vending machine for a while.
This case satisfies the rules for abandonment, but the bag is not abandoned.
Besides, the spatial rule requires high quality calibration of cameras.
Therefor, the rules for abandonment detection are modified to fit the practice application better as follows:
\begin{itemize}
\item[1)] $OP_i$ is tracked going out of the surveillance scene, i.e. its trace is extending to the edge area of given scene.
\item[2)] If $OP_i$'s trace does not reach the edge area but it disperses from the scene longer than consecutive $T = 30$ seconds and $SO_i$ is still there, then $SO_i$ is labeled as abandoned object.
\end{itemize}

\subsection{Security Event Analysis}
\label{re_id}
Three kinds of security events are considered in our work: object abandonment, moved by un-owner and stolen. They have covered most of the security events in our real lives.
The key point to recognize these security events is the verification of person and object.
Person verification is used to judge whether the people who is doing something to the under-watch object is the owner.
Object verification is to make sure that the object is moved away, or whether it present in the scene again.
Here, we use a similar approach proposed by Xiao et al.~\cite{xiao2016learning} for person and object verification. The difference is that we don't need to train a so complex CNN structure specific for person re-identification.

To reduce unnecessary computation, only the objects which have been moved and the persons who are involved in the events are verified.
When any object is being moved, the region indicated by its bounding box will be shown in the short-term foreground image $\mathbf{F}_S^t$. 
Therefore, the object whose bounding box involves foreground over a threshold of its area is counted as a possible moving object.
Then this region is cropped out from the RGB frame and is input to the FrRCNN for object detection.

If the category of the newly detected object changes, or its bounding box varies too much (over a threshold), object $SO_i$ or $BO_i$ is denoted as being moved/missed object $MO_i$.
And the person who is now closest to it is labeled as the candidate $CP_i$ for this event.
Next, $CP_i$ needs to be verified whether it is the owner of $MO_i$.
If $MO_i$ is registered in $BO_i$, $CP_i$ is labeled as suspect because the background objects belong to the surveillance scene.
If $MO_i$ is from $SO_i$, a progress of people Re-id is carried out as follows.

\paragraph*{\textbf{Owner verification}} The pose and view angle of a person influence the verification results crucially. 
For example, two pictures which are captured from a man front and rear respectively are easily identified as two different persons. 
To enhance the Re-id accuracy, 20 samples are taken for each person as follows.
When a person is labeled as the owner $OP_i$ or candidate $CP_i$, 20 frames are picked out from his first appearance till present in uniformly time interval, and 20 samples of them are cropped out from them respectively.
In this way, the appearance information of this person can be captured as different as possible.
Each sample from $CP_i$ is compared with each one from $OP_i$ using the CNN framework~\cite{xiao2016learning}. 
Then a $20 \times 20$ confused matrix is obtained to interpret the similarity of this two sets of samples. $M_{nm}$ denotes the similarity between $n$-th sample of $OP_i$ and $m$-th sample of $CP_i$. The similarity score is formally calculated as:
$S_i = \argmax_m \frac{1}{20}\sum_{n=1}^{20}M_{nm}.$
If $S_i$ is greater than a threshold, $CP_i$ and $OP_i$ are considered as the same person.  $CP_i$, $SO_i$ and $MO_i$ are canceled from the their lists respectively, because it is not necessary to pay attention on $SO_i$ any more.
Otherwise, $CP_i$ keeps the label as candidate for further watch.

In the later video frames, each newly detected object $SO_j$ is compared with each $MO_i$: $SO_j$ and $MO_i$ are cropped out from the their corresponding RGB images respectively and put into the CNN framework~\cite{xiao2016learning} to verify if $SO_j$ is $MO_i$. 
If yes, $CP_i$ is recognized as moving the object to a new place.
When $CP_i$ disperses from the surveillance scene, or it reaches a predefined regions, such as exist, $MO_i$ is not detected again. Then this event is recognized as stealing and $CP_i$ is the theft.

\begin{table}[http]
\begin{center}
\begin{tabular}{l l l l l l l }
\hline
		    & \cite{li2006evaluation} & \cite{fan2013relative} & \cite{tian2011robust} 	& \cite{lin2015abandoned} 	& ours \\
\hline
Precision  	& 0.75 	& 0.95 	& 0.85 	& 1.0 	& 1.0 \\  
Recall	   	& 1.0 	& 0.8 	& 0.8 	& 1.0 	& 1.0 \\ 
\hline
\end{tabular}
\caption{Comparison of different methods on PETS2006 video dataset.}
\label{tab:pets2006}
\end{center}
\end{table}

\begin{table*}[t!]
\centering
\begin{tabular}{|l|l|l|l|l|l|l|l|l|l|l|l|l|}
\hline
Event & \multicolumn{3}{|c|}{Abandoning} & \multicolumn{3}{|c|}{Moved by owner} & \multicolumn{3}{|c|}{Moved by un-owner} & \multicolumn{3}{|c|}{Theft} \\
\hline
Scene	&GT	&TP	&FP	&GT	&TP	&FP	&GT	&TP	&FP	&GT	&TP	&FP	\\	
\hline
Lab1-v1 & 1 & 1 & 0	& 1	& 1	& 0 & 1	& 1	& 0 & 1 & 1	& 0	\\
Lab1-v2 & 2 & 2	& 0	& 1	& 1	& 0	& 0	& 0 & 0 & 1 & 1	& 0	\\
Library & 1 & 1 & 2	& 0	& 0	& 1	& 1 & 1	& 1	& 1 & 1	& 0	\\
Lab2-v1 & 1 & 1 & 0 & 1 & 1 & 0 & 0 & 0 & 0 & 0 & 0 & 0\\
Lab2-v2 & 1 & 1 & 2 & 2 & 1 & 0 & 0 & 0 & 1 & 0 & 0 & 1\\
Lab2-v3 & 2 & 2 & 0 & 2 & 1 & 1 & 0 & 0 & 1 & 2 & 1 & 0\\
Lab2-v4 & 2 & 1 & 2 & 2 & 0 & 1 & 2 & 0 & 0 & 0 & 0 & 0\\
Hall-v1 & 0 & 0 & 0 & 0 & 0 & 0 & 1 & 1 & 0 & 1 & 1 & 0\\
Hall-v2 & 1 & 1 & 1 & 0 & 0 & 0 & 0 & 0 & 0 & 1 & 0 & 0\\
Hall-v3 & 2 & 2 & 0 & 1 & 0 & 0 & 0 & 0 & 0 & 0 & 0 & 0\\
Hall-v4 & 1 & 1 & 1 & 0 & 0 & 1 & 1 & 0 & 0 & 1 & 0 & 0\\
\hline
\hline
Sum 	&14 &13 & 8	&10	& 5 & 4 & 6 & 3 & 3 & 8 & 5 & 1\\
\hline
Precision & \multicolumn{3}{|c|}{61.9\%} & \multicolumn{3}{|c|}{55.6\%} & \multicolumn{3}{|c|}{50\%} & \multicolumn{3}{|c|}{83.3\%}\\
\hline
Recall & \multicolumn{3}{|c|}{92.8\%} & \multicolumn{3}{|c|}{50\%} & \multicolumn{3}{|c|}{50\%} & \multicolumn{3}{|c|}{62.5\%}\\
\hline
\end{tabular}
\caption{Experimental results on the video dataset SERD.}
\label{tab:serd}
\end{table*}

\subsection{CNN training}
\label{subsec:cnn}
Our neural network structure is shown as Fig.~\ref{fig:shared_cnn}. First, we use the ImageNet~\cite{russakovsky2015imagenet} pretrained CNN model for object detection.
Then the network branch for object detection is frozen and the feature network and the branch for re-identification are trained on the benchmark dataset for person re-id CUHK\cite{xiao2016learning} jointly.
The whole networks for object detection and person re-id are trained in an alternate and iterative step: each time the feature and the task networks are trained together while the other task network is frozen.
Finally, each of the task networks are fine tuned with some examples cropped from the datasets which are use in the experiments respectively.

\section{Experiments}
\label{Sec:experiment}
In this section, the performance of proposed framework is evaluated for security event recognition.
In addition, the experimental results of abandoned luggage detection will also be compared with the-state-of-the-art methods. 

The experiments are carried out on two datasets to evaluate the performance of our framework for detecting security events: abandoned object detection, recognition of objects being moved by owner or non-onwer, or stolen.
The PETS2006 is a benchmark dataset and consists of seven sequences of various scenarios. Beside the third one, each of the others includes an abandoning event.
The Security Event Recognition video dataset (SERD) is collected from four different public scene of a campus. It's constructed especially for evaluation of the proposed framework for security event recognition. It contains eleven video sequences. Each of them contains different security scenarios, such as object abandonment, theft, etc.

Our method is evaluated for detecting abandoned object on the benchmark dataset PETS2006. The experimental results are compared with the ones given by the state-of-the-art methods~\cite{li2006evaluation,fan2013relative,tian2011robust,lin2015abandoned}.
From the comparison in
Tab.~\ref{tab:pets2006} we can see that, our results are same as the one form~\cite{lin2015abandoned}, but outperforms the others. 
Furthermore, our method labels the owner of each abandoned object correctly.

We further validate the proposed method on our SERD dataset. 
Fig.~\ref{fig:stu_lab2} illustrate the whole process of a series of events about an abandoned object.
Person A comes into the lab and put his bag on the table,
and then he sits there for a long time. He is labeled as the owner of the bag, and the bag is not recognized as an abandoned object(Fig.~\ref{sugfig:come_in} and (b)).
Person B put his bag on the oscilloscope and goes out of the camera view. 
He is labeled as the owner of his bag and the bag is recognized as an abandoned object when he is going out of the scene.
Subsequently, A takes his bag away, which is recognized as allowable.
A exchanges his bag with the bag of B, which causes an alarm. 
Meanwhile, A is still labeled as the owner of his own bag.
When he is detected going out of the room, an alarm of stealing is triggered.
Each event is correctly recognized and no false alarm is triggered by our method in this video.

A summary of the experimental results is listed in Tab.~\ref{tab:serd}.
We can see that, the proposed work has similar performance on detecting events of "moved by owner" and "moved by un-owner". It means that, our method doesn't work very well by owner labeling. After analyzing the video and the experimental results we find that, the algorithm for object detection cannot provide satisfied performance: sometimes it detects objects which don't exist and cannot detect the objects of interest precisely. A better object detection methods would boost the our framework's performance. For abandoned object detection and theft recognition, our framework provides good results.


\section{Conclusion}
\label{Sec:Conclusion}
In this work, we propose a novel framework for security event recognition in surveillance videos which includes abandoned object detection and special event analysis.
It is a significant extended application of state-of-the-art works which only focus on abandoned luggage detection.
Different from previous works, our approach uses object detector, which benefits from the power of deep learning in visual tasks, instead of using foreground/background extraction for static item detection. 
The proposed approach outperforms the state-of-the-art methods for abandoned luggage detection. 
The effectiveness of our approach for more complex security event recognition has also been verified in various scenarios.
Furthermore, a new video event dataset SERD is constructed especially for the task of security event detection. SERD is collected from four different public scenes in a university campus, which contains eleven video sequences. Each of them contains different  security scenarios, such as object abandonment and theft.

%

\bibliographystyle{IEEEtran}
\bibliography{IEEEabrv,egbib}
\end{document}